\documentclass[submission,copyright,creativecommons]{eptcs}
\providecommand{\event}{Doctoral Consortium ICLP 2019} % Name of the event you are submitting to

\usepackage{breakurl}             % Not needed if you use pdflatex only.
\usepackage{underscore}           % Only needed if you use pdflatex.
\usepackage{amsmath}
\usepackage{amssymb}

\newtheorem{theorem}{Theorem} [section]
\newtheorem{lemma}{Lemma} [section]
\newtheorem{definition}{Definition} [section]

\title{Memory Management in Resource-Bounded Agents}
\author{Valentina Pitoni
\institute{L'Aquila, Italy}
\institute{DISIM, University of L'Aquila\\
}\\
\email{valentina.pitoni@graduate.univaq.it}
}

\def\titlerunning{Memory Management in Resource-Bounded Agents}
\def\authorrunning{V. Pitoni}

\begin{document}
\maketitle

\begin{abstract}
In artificial intelligence, multi agent systems constitute an interesting typology of society modeling, and have in this regard vast fields of application, which extend to the human sciences.
Logic is often used to model such kind of systems as it is easier to verify the explainability and validation, so for this reason we have tried to manage agents' memory 
extending a previous work by inserting the concept of time.
\end{abstract}

\section{Introduction}

Memory in an agent system is a process of reasoning: it is the learning process of strengthening a concept. The interaction between an agent and the environment can play an important role in constructing its memory and may affect its future behaviour. In fact, 
through memory an agent is potentially able to recall and 
to learn from experiences so that its beliefs and its future
course of action are grounded in these experiences. In computational logic, \cite{BalbianiDL16} introduces DLEK (Dynamic Logic of Explicit beliefs and Knowledge) as a logical formalization of the short-term and long-term memory. The underlying idea is to represent reasoning about the formation of beliefs through perception and inference in non-omniscient resource-bounded agents. DLEK has however no notion of time, while agents' actual perceptions are inherently timed and so are many of the inferences drawn from such perceptions.
In this paper we present an extension of LEK/DLEK to T-LEK/T-DLEK (``Timed LEK'' and ``Timed DLEK'') obtained by introducing a special function which associates to each belief the arrival time and controls timed inferences.
Through this function it is easier to keep the evolution of the surrounding world under control and the representation is more complete. This abstract is an evolution version of \cite{CostantiniFP18}, where we have introduced explicit time instants and time intervals in formulas, and it is
extracted from \cite{Costantini}.

\section{T-LEK and T-DLEK}\label{tdlek}
As in \cite{BalbianiDL16}, our logic consists of two different components: a static component, called T-LEK, which is mix between an Epistemic Logic and Metric Temporal Logic, and a dynamic component, called T-DLEK, which extends the static one with mental operations, which are vary important for ``controlling'' beliefs (adds new belief, update belief, etc).

\subsection{Syntax}
In our scenario we fix $Atm=\{p(t_1,t_2),q(t_3,t_4) \mbox{, ... ,} h(t_i,t_j) \}$ where $t_i \leqslant t_j $ and $p,q,h$ are predicates, that can be equal or not. 
Moreover $p(t_1,t_2)$ stands for \textit{``p is true from the time instant $t_1$ to $t_2$"} with $t_1, t_2 \in \mathbb{N}$ (\textit{Temporal Representation} of the external world); 
as a special case we can have $p(t_1,t_1)$ which stands for \textit{``p is true in the time instant $t_1$"}. Obviously we can have predicates with more terms than only two but in that case we fix that the first two must be those that identify the time duration of the belief (i.e. $ open(1,3,\mbox{door})$ which means ``the agent knows that the door is open from time 1 to time 3''). Instead in the previous work \cite{CostantiniFP18} we considered atoms of the form $p_I$ with $I = [t_1,t_2]$, which are the conjunction $p_{t_1} \wedge\ p_{t_1+1} \wedge\cdots\wedge\ p_{t_2}$
and also $p_t$ stand for $p_{I_t}$ with $I_t=[t,t]$; we have decided to change approach because $p_I$ is too detached from propositional logic.

Below is the definition of the formulas of the language ${\mathcal L}_{T\mbox{-}LEK}$, with a slight abuse, in this grammar we use  $I$ as  terminal symbol
standing for  time intervals (possibly specified through arithmetic expressions, as said earlier):
\[\begin{array}{rcl}
\varphi, \psi              & ~:=~ &  p(t_1,t_2)\ |\ \neg \varphi\ |\ \square_I\, \varphi\ |\ B\,\varphi\ |\ K\,\varphi |\ \varphi\ \wedge\ \psi  |\ \varphi\ \rightarrow\ \psi \\
\end{array}\]
Other Boolean connectives $\top$, $\bot$, $\leftrightarrow$ are defined from $\neg$ and  $\wedge$ as usual.
In the formula $\square_{I}\, \Phi$ the MTL Interval ``always'' operator is applied to a formula;
$I$ is a ``time-interval'' which is a closed finite interval $[t,l]$ or an infinite interval $[t,\infty)$ (considered open on the upper bound),
for any expressions/values $t,l$ such that $0\leq t \leq l$
and $\square_{[0,\infty)}$ will sometimes be written simply as~$\square$.
The operator $B$ is intended to denote belief and the operator $K$ to denote knowledge. 
More precisely, $B$ identifies beliefs present in the working memory, instead $K$ identifies what rules present in the background knowledge.

Terms/atoms/formulas as defined so far are \emph{ground}, namely there are no variables occurring therein.
We introduce variables and use them in formulas in a restricted manner, as usual for example in answer set programming.
Variables can occur in formulas in any place; constants can occur and are intended as place holders for elements of the Herbrand universe.
More specifically, a ground \emph{instance} of a term/atom/formula involving variables is obtained by uniformly substituting ground terms to all variables (\emph{grounding} step),
with the restriction that any variable occurring in an arithmetic expression (i.e., specifying a time instant) can be replaced by a (ground) arithmetic expressions only.
Consequently, a non-ground term/atom/formula represents the possibly infinite set of its ground instances, namely, its \emph{grounding}. Notice that the rational of considering ground  formulas is that they represent perceptions (either new or already recorded in agent's memory) coming in general from the external world (we say ``in general'' as, in fact, in some of the aforementioned agent-oriented frameworks perceptions can also result from \emph{internal events},
i.e., from an agent's observations of its own internal activities).
As it is customary in logic programming, variable symbols are indicated with an initial uppercase letter 
whereas constants/functions/predicates symbols are indicated with an initial lowercase letter.

The language ${\mathcal L}_{T\mbox{-}DLEK}$ of Temporalized DLEK (T-DLEK) is obtained
by augmenting ${\mathcal L}_{T{-}LEK}$ with the expression $[\alpha]\,\psi$, where $\alpha$ denotes a \emph{mental operation} and 
$\psi$ is a ground formula.
The mental operations that we consider are essentially the same as in \cite{BalbianiDL16}:
\begin{itemize}
	\item
	$+ \varphi$, where $\varphi$ is a ground formula of the form $p(t_1,t_2)$ or $\neg p(t_1,t_2)$: the mental operation that serves to form a new belief from a perception~$\varphi$.
	A perception may become a belief whenever an agent becomes ``aware'' of the perception and takes it into explicit consideration.
	Notice that $\varphi$ may be a negated atom.
	
	\item
	$\cap(\varphi,\psi)$: believing both $\varphi$ and $\psi$, an agent starts believing their conjunction.
	
	\item
	${\vdash}(\varphi,\psi)$, where $\psi$ is a ground atom, say~$p(t_1,t_2)$:
	an agent, believing that $\varphi$ is true and having in its long-term memory
	that $\varphi$ implies $\psi$ (in some suitable time interval including~$[t_1,t_2]$),
	starts believing that $p(t_1,t_2)$ is true.
	
	\item
	${\dashv}(\varphi,\psi)$  where $\varphi$ and $\psi$ are ground atoms, say $p(t_1,t_2)$ and $q(t_3,t_4)$ respectively:
	an agent, believing $p(t_1,t_2)$ and having in the long-term memory
	that $p(t_1,t_2)$ implies $\neg q(t_3,t_4)$, removes the timed belief $q(t_3,t_4)$
	if the intervals match. Notice that, should $q$ be believed in a wider interval I such that $[t_1,t_2] \subseteq I$, the belief $q(.,.)$ is removed concerning intervals $[t_1,t_2]$ and $[t_3,t_4]$, but it is left for the remaining sub-intervals (so, its is ``restructured'').
\end{itemize}
The last mental operation, which is a sort of ``update" or 
``restructuring operator", substitutes $- \varphi$ (\cite{BalbianiDL16}), that instead represents arbitrary ``forgetting'', i.e., removing  a belief from the short-term memory. In fact in \cite{BalbianiDL16} there are $+ \varphi$, $- \varphi$, ${\vdash}(\varphi,\psi)$ and ${\dashv}(\varphi,\psi)$.

\noindent
\textbf{Example 1:} We propose a small example to illustrate the form and the role of rules in the working memory and in the long-term memory. If at time $t{=}2$ it is starting raining, in the agent's working memory there will be the following belief: $B (raining(2,2))$. And if we have in the background knowledge $K(rain(t_1,t_2) \rightarrow take(t_1,t_2, \mbox{umbrella}) )$ and $2 \in [t_1,t_2]$ then the agent can infer $B(take(2,2,\mbox{umbrella}))$, which is a new belief stored in the working memory. And if we have also $K(rain(t_1,t_2) \wedge $ \\$take(t_1,t_2, \mbox{umbrella}) \rightarrow go(t_1+1,\infty,\mbox{shops}) )$ than the agent can infer $B(go(3,\infty,\mbox{shops}))$ which means that after getting the umbrella the agent can go around the shops.\\

\noindent
\textbf{Example 2:} An example of a non-ground T-LEK formula is:\\
$$K( \square_{[t_1,t_2]}(\mathit{enrollment(T,T))} \ \rightarrow \square_{[t_1,t_2]}(\square_{[T,T+14]} \mathit{send\_payment(T_1,T_1)}))$$
where we suppose that an agent knows that it is possible to enroll in the university in the period $[t_1,t_2]$ and that,
after the enrollment, the payment must be sent within fourteen days (still staying within the interval $[t_1,t_2]$).
Since, by the restrictions on formulas stated earlier, it must be the case that
$T_1 \in [T,T+14]$ and both $T$, $T+14$ must be in $[t_1,t_2]$, only a finite set of ground instances
of this formula can be formed by substituting natural numbers to the variables~$T,T_1$ 
(specifically, the maximum number of ground instances is $t_2-t_1-14+1$ assuming to pay on the last day $t_2$). 
In case one would consider the more general formula:
$$K( \square_{[t_1,t_2]}(\mathit{enrollment(T,T,X))} \ \rightarrow \square_{[t_1,t_2]}(\square_{[T,T+14]} \mathit{send\_payment(T_1,T_1,X)}))$$
where $X$ represents a student of that university, i.e., $\mathit{student(.,.,X)}$ holds for some ground instance of~$X$,
then the set of ground instances would grow, as a different instance should be generated for each student (i.e., for each ground term replacing~$X$).
In practice, however, ground instances need not to be formed a priori, but rather they can be generated upon need when applying a rule;
in the example, just one ground instance should be generated when some student intends to enroll
in that university at a certain time $T=\hat{t}$.

\subsection{Semantics}
Semantics of DLEK and T-DLEK are both based on a set $W$ of worlds.
In both DLEK and T-DLEK we have the valuation function: $V : W \rightarrow 2^{\mathit{Atm}}$. Also we define the ``time'' function $T$ that associates to each formula the time interval in which this formula is true and operates as follows:
\begin{itemize}
	\item $T(p(t_1,t_2))= [t_1,t_2]$, which stands for \textit{``p is true in the time interval $[t_1,t_2]$"} where $t_1,t_2 \in \mathbb{N}$; as a special case we have $T(p(t_1,t_1))= t_1$,  which stands for \textit{``p is true in the time instant $t_1$"} where $t_1 \in \mathbb{N}$ (time instant);
	\item $T(\neg p(t_1,t_2))= T(p(t_1,t_2))$, which stands for \textit{``p is not true in the time interval $[t_1,t_2]$"} where $t_1,t_2 \in \mathbb{N}$;
	\item $T(\varphi \mbox{ op } \psi)= T(\varphi) \biguplus T(\psi)$ with ${op}\in\{\vee,\wedge,\rightarrow\}$, which means the unique smallest interval including both $T(\varphi)$ and $T(\psi)$;
	\item $T(B \varphi)= T(\varphi)$;
	\item $T(K \varphi)= T(\varphi)$;
	\item $T(\square_{I} \varphi)= I$ where $I$ is a time interval in $\mathbb{N}$;
	\item $T([\alpha]\varphi)$ there are different cases depends on which kind of mental operations we applied: 
	\begin{enumerate}
		\item $T(+ \varphi)=T(\varphi)$;
		\item $T(\cap(\varphi,\psi)) = T(\varphi) \biguplus T(\psi)$;
		\item $T({\vdash}(\varphi,\psi))= T(\psi) $;
		\item $T({\dashv}(\varphi,\psi))$ returns the restored interval where $\psi$ is true.
	\end{enumerate}
\end{itemize}
For a world $w$, let $t_1$ be the minimum time instant of $T(p(t_1,t_1))$ where $p(t_1,t_1) \in V(w)$  and let $t_2$ be the supremum time instant (we can have $t_2 =\infty$) among the atoms in~$V(w)$. Then, whenever useful, we denote $w$ as $w_I$ where $I = [t_1,t_2]$, which
identifies the world in a given interval.

The notion of LEK/T-LEK model does not consider mental operations, discussed later, and is introduced by the following definition.

\begin{definition}
	A \emph{ T-LEK model} is a tuple $M=\langle W; N; R; V; T \rangle$ where:
	\begin{itemize}
		\item $W$ is the set of worlds;
		\item $V : W \rightarrow 2^{\mathit{Atm}} $ valuation function;
		\item $T$ ``time'' function;
		\item $R \subseteq W{\times}W$ is the accessibility relation, required to be an equivalence relation so as to model omniscience in the background knowledge s.t. $R(w)=\lbrace v \in W \mid w_I R\ v_{I}\rbrace$ called \emph{epistemic state of the agent in $w_{I}$}, which indicates all the situations that the agent considers possible in the world $w_I$ or, equivalently any situation the agent can retrieve from long-term memory based on what it knows in world $w_I$;
		\item $N : W \rightarrow 2^{2^W}$ is a \emph{``neighbourhood''} function,  $ \forall w \in W$, $N(w)$ defines, in terms of sets of worlds, what the agent is allowed to explicitly \emph{believe} in the world $w_I$;  $ \forall w_{I}, v_{I} \in W$, and $X \subseteq W$:
		\begin{enumerate}
			\item
			if $ X \in N(w_I)$, then $X \subseteq R(w_{I})$: each element of the neighbourhood is a set composed of reachable worlds;
			\item
			if $w_{I} R\ v_{I}$, then $N(w_I) \subseteq N(v_{I})$: if the world $v_{I}$ is compliant with the epistemic state of world $w_I$, then the agent in the world $w_I$ should have a subset of beliefs of the world $v_{I}$.
		\end{enumerate}
	\end{itemize}
\end{definition}

A preliminary definition before the Truth conditions : let $M = \langle W; N; R; V; T \rangle$ a T-LEK model. Given a formula $\varphi$, for every  $w_I \in W$,\,
we define $$\parallel \varphi \parallel^M_{w_I} = \{v_I \in W \mid M,v_I \models \varphi\} \cap R(w_I).$$
Truth conditions for T-DLEK formulas are defined inductively as follows:
\begin{itemize}
	\item $M, w_I \models p(t_1,t_2)$ iff $p(t_1,t_2) \in V(w_I)$ and $T(p(t_1,t_2)) \subseteq I$;
	\item $M, w_I \models \neg \varphi $ iff $M, w_I \nvDash \varphi $ and $T(\neg \varphi)\subseteq I$;
	\item $M, w_I \models \varphi \wedge \psi $ iff $M, w_I \models \varphi $ and $M, w_I \models \psi $ with $T(\varphi),T(\psi) \subseteq I$;
	\item $M, w_I \models \varphi \vee \psi $ iff $M, w_I \models \varphi $ or $M, w_I \models \psi $ with $T(\varphi),T(\psi) \subseteq I$;
	\item $M, w_I \models \varphi \rightarrow \psi $ iff $M, w_I \nvDash \varphi $ or $M, w_I \models \psi $ with $T(\varphi),T(\psi) \subseteq I$;
	\item $M, w_I \models  B\, \varphi$ iff \,$\parallel\varphi\parallel^M_{w_I} \in N(w_I)$ and $T(\varphi) \subseteq I$;
	\item $M, w_I \models K_i\, \varphi$ iff for all $v_{I} \in R(w_I)$, it holds that $M, v_{I} \models \varphi$ and $T(\varphi) \subseteq I$;
	\item  $M, w_I \models \square_{J} \varphi$  iff $T(\varphi) \subseteq J \subseteq I$ and  for all $v_{I} \in R(w_I)$, it holds that  $M, v_{I} \models \varphi$;
\end{itemize}

In particular, considering formulas of the forms  $B\, \varphi$ and $K\, \varphi$, we observe that $M, w_I \models B\, \varphi$
if the set $\parallel \varphi\parallel^M_{w_I}$ of worlds reachable from $w_I$ which entail $\varphi$ in the very same model $M$
belongs to the \emph{neighbourhood} $N(w_I)$ of $w_I$.
Hence, knowledge pertains to formulas entailed in model $M$ in every reachable world,
while beliefs pertain to formulas entailed only in some set of them,
where this set must however belong to the neighbourhood and so it must be composed of reachable worlds.
Thus, an agent is seen as omniscient with respect to knowledge, but not with respect to beliefs.

Concerning a mental operation $\alpha$ performed by any agent~$i$, we have: 
$M, w_I \models [\alpha]\,\varphi$ iff $M^{\alpha}, w_I \models \varphi$ and $T(\varphi) \subseteq I$ where
$M^{\alpha} = \langle W; N^{\alpha}(w_I); R  ; V; T \rangle$
Here  $\alpha$ represents a mental operation affecting the sets of beliefs. 
In particular, such operation can add new beliefs by direct perception, by means of one inference step,  or as a conjunction of previous beliefs.
When introducing new beliefs, the neighbourhood must be extended accordingly, as seen below; in particular,
the new neighbourhood $N^{\alpha}(                                                                                                                             w_I)$ is defined for each of the mental operations as follows.
\begin{itemize}
	
	\item Learning perceived belief:
	
	$N^{+ \varphi}(w_I) = N(w_I)\, \cup \big\{\parallel \varphi \parallel_{w_I}^M\big\}$ with $T(\varphi) \subseteq I$.
	
	The agent adds to its beliefs perception $\varphi $ (namely, an atom or the negation of an atom) perceived at a time in $T(\varphi)$;
	the neighbourhood is expanded to as to include the set composed of all the reachable worlds which entail $\varphi$ in~$M$.
	
	\item Beliefs conjunction:
	
	$N^{\cap(\psi,\chi)}(w_I)=
	\left\{\begin{array}{ll}
	N(w_I) \,\cup \big\{\parallel \psi \wedge \chi \parallel_{w_I}^M\big\} & \mbox{if } M,{w_I} \models B(\psi) \wedge B(\chi) \\ 
	&  \mbox{and } T(\cap(\psi,\chi)) \subseteq I \\
	N(i,w_I)  & \mbox{otherwise}
	\end{array}\right.$
	
	The agent adds $\psi \wedge \chi$ as a belief if it has among its previous beliefs
	both $\psi$ and $\chi$, with $I$ including all time instants referred to by them;
	otherwise the set of beliefs remain unchanged.
	The neighbourhood is expanded, if the operation succeeds, with those sets of reachable worlds where both formulas are entailed in~$M$.
	
	\item Belief inference:
	
	$N^{\vdash(\psi,\chi)}(w_I)= 
	\left\{\begin{array}{ll}
	N(w_I) \,\cup \big\{\parallel \chi \parallel_{w_I}^M\big\} & \mbox{if } M,{w_I} \models B(\psi)\ \wedge\ K(\psi\rightarrow\chi) \\ 
	&  \mbox{and } T(\vdash(\psi,\chi)) \subseteq I \\
	N(w_I) & \mbox{otherwise}
	\end{array}\right.$
	
	The agent adds the ground atom $\chi$ as a belief in its short-term memory if it has $\psi$ among its previous beliefs
	and has in its background knowledge $K(\psi \rightarrow \chi)$,
	where all the time stamps occurring in $\psi$  and in $\chi$ belong to~$I$.
	Observe that, if~$I$  does not include all time instants involved in the formulas, 
	the operation does not succeed and thus the set of beliefs remains unchanged.
	If the operation succeeds then the neighbourhood is modified by adding $\chi$ as a new belief.

	\item Beliefs revision (applied only on ground atoms):\\ Given $Q=q(j,k)$ s.t. $T(q(j,k))=$ $T(q(t_1,t_2)) \cap T(q(t_3,t_4))$ with $j,k \in \mathbb{N}$ and $P=\big\{$$ M,{w_I} \models  B(p(t_1,t_2)) \wedge B(q(t_3,t_4)) \wedge K(p(t_1,t_2) \rightarrow \neg q(t_3,t_4))$ and $T(\dashv(p(t_1,t_2),q(t_3,t_4))) \subseteq I$ and there is no interval $J \supsetneq T(p(t_1,t_2))$ s.t. $B(q(t_5,t_6))$ where $T(q(t_5,t_6)){=}J\big\}$:
	
	$N^{\dashv(p(t_1,t_2),q(t_3,t_4))}(w_I)=
	\left\{\begin{array}{ll}
	N(w_I) \setminus \big\{\parallel Q \parallel_{w_I}^M\big\} & \mbox{if P}\\
	N(i,w_I)  & \mbox{otherwise}
	\end{array}\right.$

	The agent believes that $q(t_3,t_4)$ holds only in the interval $T(q(t_3,t_4))$ and has the perception of $p(t_1,t_2)$ where $T(p(t_1,t_2))\subseteq T(q(t_3,t_4))$. 
	Then, the agent replaces previous belief $q(t_3,t_4)$ in the short-term memory with $q(t_5,t_6)$ where $T(q(t_5,t_6)){=}T(q(t_3,t_4)) \setminus T(q(t_1,t_2))$. 
	In general, the set $T(q(t_3,t_4)) \setminus T(q(t_1,t_2))$ is not necessarily an interval:
	being $T(p(t_1,t_2))\subseteq T(q(t_3,t_4))$, with $T(p(t_1,t_2)){=}[t_1, t_2]$, and $T(q(t_3,t_4)){=}[t_3,t_4]$, we have that $T(q(t_3,t_4)) \setminus T(q(t_1,t_2)){=} [t_3,t_1-1]{\cup}[t_2+1,t_4]$.
	Thus, $q(t_3,t_4)$ is replaced by $q(t_3,t_1-1)$ and $q(t_2+1,t_4)$ (and similarly if $t_4 = \infty$).
	
\end{itemize}
We write $\models_{{T\mbox{-}DLEK}} \varphi$ to denote that $\varphi$ 
is true in all worlds $w_I$, of every TLEK model~$M$.\\

\noindent
\textbf{Example 3:}
Let us consider the example of a person who is married or divorced, where s(he) can perform the \emph{action} to be married or divorced.
Let us assume that performed actions are recorded among an agent's perceptions, with the due time stamp.
For reader's convenience, actions are denoted using a suffix ``$A$".
For simplicity, actions are supposed to always succeed and to produce an effect within one time instant.
Let us consider the following rules (kept in long-term memory):
\[\begin{array}{ll}
K(\mathit{marry(T,T)A} \rightarrow \mathit{married(T+1,\infty)}) \\
K (\mathit{divorce(T,T)A} \rightarrow   \mathit{divorced(T+1,\infty)}).
\end{array}\]
Let us now assume that a person married, e.g., at time~$5$; then, a belief will be formed of the person is married from time~$6$ on; however, if that person later divorced, e.g., at time~$8$, as a consequence result that s(he) is divorced from time~$9$.
It can be seen that the application of previous rules in consequence of an agent's action of marring/divorcing determines some ``belief restructuring'' in the short-term memory of the agent.
In absence of other rules concerning marriage, we intend that a person can not be simultaneously married and divorced.
The related belief update is determined by the following rules:
\[\begin{array}{ll}
K (\mathit{married(T,\infty)} \rightarrow \neg \mathit{divorced(T,\infty}))  \\
K (\mathit{divorced(T,\infty)} \rightarrow \neg \mathit{married(T,\infty)})
\end{array}\]
With the above timing, the result of their application is that the belief formed at time~$5$,
i.e., ${married(6,\infty)}$ will be replaced by ${married(6,8)}$ plus ${divorced(9,\infty)}$.\\

\noindent
\textbf{Property 1:}
For the mental operations previously considered we have the following
(where $\varphi,\psi$ are as explained earlier):
\begin{itemize}
	\item
	$\models_{{T\mbox{-}DLEK}} [+ \varphi]B \varphi$.\\
	Namely, as a consequence of the operation $+\varphi$ (thus after the perception of  $\varphi$)
	the agent $i$ adds $\varphi$ to its beliefs.
	
	\item
	$\models_{{T\mbox{-}DLEK}} (B\varphi \wedge B\psi) \rightarrow [\cap(\varphi,\psi)]B(\varphi \wedge \psi)$.\\
	Namely, if an agent has $\varphi$ and $\psi$ as beliefs, then as a consequence of the mental operation $\cap(\varphi,\psi)$
	the agent starts believing $\varphi \wedge \psi$;
	
	\item 
	$\models_{{T\mbox{-}DLEK}} (K(\varphi \rightarrow \psi ) \wedge B\,\varphi)
	\rightarrow
	[{\vdash}(\varphi,\psi)]\,B\,\psi$.\\
	Namely, if an agent has $\varphi$ as one of its beliefs and has $K(\varphi{\rightarrow}\psi)$ in its background knowledge,
	then as a consequence of the mental operation ${\vdash}(\varphi,\psi)$ the agent starts believing~$\psi$;
	
	\item 
	$\models_{{T\mbox{-}DLEK}} (K(p(t_1,t_2) \rightarrow \neg q(t_3,t_4)) \wedge B\,(p(t_1,t_2)) \wedge B\,(q(t_3,t_4))) \rightarrow
	[{\dashv}(p(t_1,t_2), q(t_3,t_4))]\,B\, (q(t_5,t_6))$\\ where $T(q(t_5,t_6)) = T(q(t_3,t_4)) \setminus T(q(t_1,t_2))$.\\
	Namely, if an agent has $q(t_3,t_4)$ as one of its beliefs, $q$ is not believed outside $T(q(t_3,t_4))$, the agent perceives $p(t_1,t_2)$ where $T(p(t_1,t_2))\subseteq T(q(t_3,t_4))$, and has $K(p(t_1,t_2) \rightarrow \neg q(t_3,t_4))$ in its background knowledge. Then after the mental operation ${\dashv}(p(t_1,t_2),q(t_3,t_4))$ the agent starts believing $q(t_5,t_6))$ where $T(q(t_5,t_6)) = T(q(t_3,t_4)) \setminus T(q(t_1,t_2))$.
	
\end{itemize}

\section{Axiomatization and Canonical Models}
\label{axiomsmodel}

The logic T-DLEK can be axiomatized as an extension of the axiomatization of DLEK as follows.
We implicitly assume modus ponens, standard axioms for classical propositional logic, and the necessitation rule.
The T-LEK axioms are the following:
\begin{enumerate}
	\item $K(\varphi)\wedge K(\varphi\rightarrow\psi) \rightarrow K(\psi)$;
	\item $K(\varphi)\rightarrow \varphi$;
	\item $K(\varphi)\rightarrow K K(\varphi)$;
	\item $\neg K(\varphi)\rightarrow K\neg K(\varphi)$;
	\item $B\varphi\wedge K(\varphi\leftrightarrow\psi )\rightarrow B\psi$. 
\end{enumerate}

The axiomatization of T-DLEK, involves these axioms:
\begin{enumerate}
	\item $[\alpha] f \leftrightarrow f$ where $f = p$ or $f = p_t$ or $f = p_{I}$;
	\item $[\alpha] \neg \varphi \leftrightarrow \neg [\alpha]\varphi$;
	\item $[\alpha](\varphi \wedge \psi) \leftrightarrow [\alpha]\varphi \wedge [\alpha]\psi$;
	
	\item $[\alpha] K (\varphi)\leftrightarrow K\big([\alpha](\varphi)\big)$;
	
	\item $[+\varphi]B\psi\leftrightarrow\Big(B([+\varphi]\psi)\vee K\big([+\varphi]\psi\leftrightarrow\varphi\big)\Big)$;	
	
	\item $[{\vdash}(\varphi,\psi)]B\chi\leftrightarrow \Big(B\big([{\vdash}(\varphi,\psi)]\chi\big) \vee
	\Big(B\varphi\wedge K\big(\varphi\rightarrow\psi\big) \wedge K \big([{\dashv}(\varphi,\psi)]\chi\leftrightarrow\psi\big)\Big)\Big)$;
	
	\item $[{\dashv}(\varphi,\psi)]B\chi\leftrightarrow\Big(B\big([{\dashv}(\varphi,\psi)]\chi\big)\vee
	\Big(B\varphi\wedge K(\varphi{\rightarrow}\neg\psi)\wedge K \big([{\dashv}(\varphi,\psi)]\chi{\leftrightarrow}{\neg}\psi\big)\Big)\Big)$;
	
	\item $[\cap(\varphi,\psi)]B\chi\leftrightarrow\Big(B\big([\cap(\varphi,\psi)]\chi\big)\vee\Big((B \varphi \wedge B \psi) \wedge \ K \big([\cap(\varphi,\psi)]\chi\leftrightarrow (\varphi \wedge \psi)\big)\Big)$;
	
	\item $\displaystyle\frac{\psi\leftrightarrow\chi}{\varphi\leftrightarrow\varphi[\psi/\chi]}$~~
	where $\varphi[\psi/\chi]$ denotes the formula obtained by replacing~$\psi$ with~$\chi$ in~$\varphi$. 
\end{enumerate}

We write ${\rm T\mbox{-}DLEK}\vdash\varphi$ to indicate that  $\varphi$ is a theorem of TDLEK.

Both logics T-LEK and T-DLEK are sound for the class of T-LEK models.
The proof that T-DLEK is strongly complete can be achieved by using a standard canonical model argument.

The \emph{canonical T-LEK model} is a tuple $M_c = \langle W_c; N_c; $ $\lbrace R_c  ; V_c; T_c \rangle$ where:
\begin{itemize}
	\item $W_{c}$ is the set of all maximal consistent subsets of $\mathcal{L}_{T\mbox{-}LEK}$; so, as in \cite{BalbianiDL16}, canonical models are constructed from worlds which are sets of syntactically correct formulas of the underlying language and are in particular the largest consistent ones.
	As before, each $w \in W_{c}$ can be conveniently indicated as $w_I$.
	
	\item For every $w_{I} \in W$  and $w_{I} R_{c} v_{I}$ if and only if
	$K \varphi\in w_{I}$ iff $K \varphi\in v_{I}$;
	i.e., $R_{c}$ is an equivalence relation on knowledge; as before,
	we define
	$R_{c}(w_{I}) = \lbrace v_{I}\in W\mid w_{I} R_{c_i} v_{I}\rbrace$. 
	Thus, we cope with our extension from knowledge of formulas to knowledge of formulas.
	
	\item  Analogously to \cite{BalbianiDL16}, for $w_{I} \in W$, $ \Phi\in \mathcal{L}_{T\mbox{-}LEK}$  we define 
	$A_{\Phi}(w_{I})= \lbrace v_{I} \in R_{c}(w_{I}) \mid \Phi\in v_{I}\rbrace$.
	Then, we put
	$N_c(w_{I})= \lbrace A_{\Phi}(w_{I}) \mid B \Phi\in w_{I} \rbrace$.
	\item $V_c$ is a valuation function defined as before.
	\item $T_c$ is a ``time'' function defined as before.
\end{itemize}

As stated in Lemma 2 of \cite{BalbianiDL16}, 
there are the following immediate consequences of
the above definition: if $w_{I} \in W_c$ and $i \in \mathit{Ag}$, then 
\begin{itemize}
	\item
	for $\Phi\in\mathcal{L}_{T\mbox{-}LEK}$, it holds that 
	$K \Phi\in w_{I}$ if and only if $\forall v_{I}\in W$ such that $w_{I}R_{c}v_{I}$
	we have $\Phi\in v_{I}$;
	\item
	for $\Phi\in\mathcal{L}_{T\mbox{-}LEK}$, if $B \Phi\in w_{I}$ and $w_{I}R_{c}v_{I}$ then $B \Phi\in v_{I}$.
\end{itemize}

Thus, while $R_{c}$-related worlds have the same knowledge and $N_c$-related worlds have the same beliefs, as stated in Lemma 3 of \cite{BalbianiDL16} there can be $R_{c}$-related worlds with different beliefs. 
The above properties can be used analogously to what is done in \cite{BalbianiDL16} to prove that, by construction, the following results hold:

\begin{lemma}
	For all $w_I \in W_c$ and $B_i\Phi, B_i\Psi\in\mathcal{L}_{T\mbox{-}LEK}$, if $B_i\Phi\in w_I$  but $B_i\Psi\not\in w_{I}$,
	it follows that there exists $v_{I}\in R_{c_i}(w_{I})$ such that
	$\Phi\in v_{I} ~\leftrightarrow~\Psi\not\in v_{I}$.
	
\end{lemma}

\begin{lemma}
	For all $\Phi\in\mathcal{L}_{T\mbox{-}LEK}$ and $w_I\in W_c$ it holds that $\Phi\in w_I$ if and only if
	$M_c,w_I\vDash\Phi$.
\end{lemma}

\begin{lemma}
	For all $\Phi\in\mathcal{L}_{T\mbox{-}DLEK}$ then there exists $\tilde{\Phi}\in\mathcal{L}_{T\mbox{-}LEK}$ such that ${\rm T\mbox{-}DLEK}\vdash\Phi\leftrightarrow\tilde{\Phi}$.
\end{lemma}

\noindent
Under the assumption that the interval $I$ is finite,
the previous lemmas allow us to prove the following theorems.
The limitation to finite intervals is not related to features of the proposed approach,
but to well-known paradoxes of temporal logics on infinite intervals.

\begin{theorem}
	{T-LEK} is strongly complete for the class of \emph{T-LEK} models.
\end{theorem}

\begin{theorem}
	{T-DLEK} is strongly complete for the class of \emph{T-LEK} models.
\end{theorem}

With the new formalization of time intervals proposed in this paper, the proof of the previous Theorem immediately follows from the proof proposed in \cite{BalbianiDL16}.

\section{Conclusion}\label{conclusions}
In this work we extended an existing approach to the logical modeling of short-term and long-term memories in Intelligent Resource-Bounded Agents by introducing the $T$ function, which manages the interval when an atom is true. Through this function we are also able to assign a ``timing" to the epistemic operators $B$ and $K$. Moreover we add the always operator $\square_{I}$ of the Metric Temporal Logic to increase the expressiveness of our logic. We considered not just adding new beliefs, rather we introduced a new mental operation not provided in DLEK, to allow for removing/restructuring existing beliefs. The resulting T-DLEK logic shares similarities in the underlying principles with hybrid logics (cf., e.g., \cite{Hylogic}) and with temporal epistemic logic (cf., e.g., \cite{TEL}); as concerns the differences, the former has time instants but no time intervals, and the latter has neither time instants nor time intervals.

With regard to complexity for the mono agent case for LEK it has been proved that the satisfiability problem is decidable and it has been proved to be in NP-complete, instead for DLEK it has been conjectured to be PSPACE. It is easy to believe that our extensions cannot spoil decidability because the $T$ function do not interfere. Inference steps to derive new beliefs are analogous to D-LEK: just one modal rule at a time is used and a sharp separation is postulated between the working memory, where inference is performed, and the long-term memory.
We are working on the extension to the multi-agent case, also reconsidering the complexity, and on the ``Store'' operation, which consist in managing the transition from working memory to long term memory.

\nocite{*}
\bibliographystyle{eptcs}

\end{document}